\documentclass[10pt,twocolumn,letterpaper]{article}
\pdfoutput=1

\usepackage{cvpr}
\usepackage{times}
\usepackage{epsfig}
\usepackage{graphicx}
\usepackage{amsmath}
\usepackage{amssymb}
\usepackage{caption}
\usepackage{subcaption}
\usepackage{diagbox}
\usepackage{verbatim}
\usepackage{pdfpages}

\newcommand{\RNum}[1]{\uppercase\expandafter{\romannumeral #1\relax}}

\newcommand\blfootnote[1]{%
  \begingroup
  \renewcommand\thefootnote{}\footnote{#1}%
  \addtocounter{footnote}{-1}%
  \endgroup
}

\usepackage[breaklinks=true,bookmarks=false]{hyperref}

\cvprfinalcopy 

\def\cvprPaperID{1694} 

\ifcvprfinal\fi
\begin{document}

\title{Spatiotemporal Fusion in 3D CNNs: A Probabilistic View}

\author{Yizhou Zhou$^{*1}$
\and
Xiaoyan Sun$^{\dagger 2}$
\and
Chong Luo$^{2}$
\and
Zheng-Jun Zha$^{1}$
\and
Wenjun Zeng$^{2}$
\and
$^{1}$University of Science and Technology of China\\
{\tt\small {zyz0205@mail.ustc.edu.cn,  zhazj@ustc.edu.cn}}
\and
$^{2}$Microsoft Research Asia\\
{\tt\small {\{xysun,cluo,wezeng\}}@microsoft.com}
}

\maketitle
\thispagestyle{empty}

\begin{abstract}
  Despite the success in still image recognition, deep neural networks for spatiotemporal signal tasks (such as human action recognition in videos) still suffers from low efficacy and inefficiency over the past years. Recently, human experts have put more efforts into analyzing the importance of different components in 3D convolutional neural networks (3D CNNs) to design more powerful spatiotemporal learning backbones. Among many others, spatiotemporal fusion is one of the essentials. It controls how spatial and temporal signals are extracted at each layer during inference. Previous attempts usually start by ad-hoc designs that empirically combine certain convolutions and then draw conclusions based on the performance obtained by training the corresponding networks. These methods only support network-level analysis on limited number of fusion strategies. In this paper, we propose to convert the spatiotemporal fusion strategies into a probability space, which allows us to perform network-level evaluations of various fusion strategies without having to train them separately. Besides, we can also obtain fine-grained numerical information such as layer-level preference on spatiotemporal fusion within the probability space. Our approach greatly boosts the efficiency of analyzing spatiotemporal fusion. Based on the probability space, we further generate new fusion strategies which achieve the state-of-the-art performance on four well-known action recognition datasets.
\end{abstract}

\section{Introduction}
\begin{figure}
    \centering
    \includegraphics[width=0.99\linewidth]{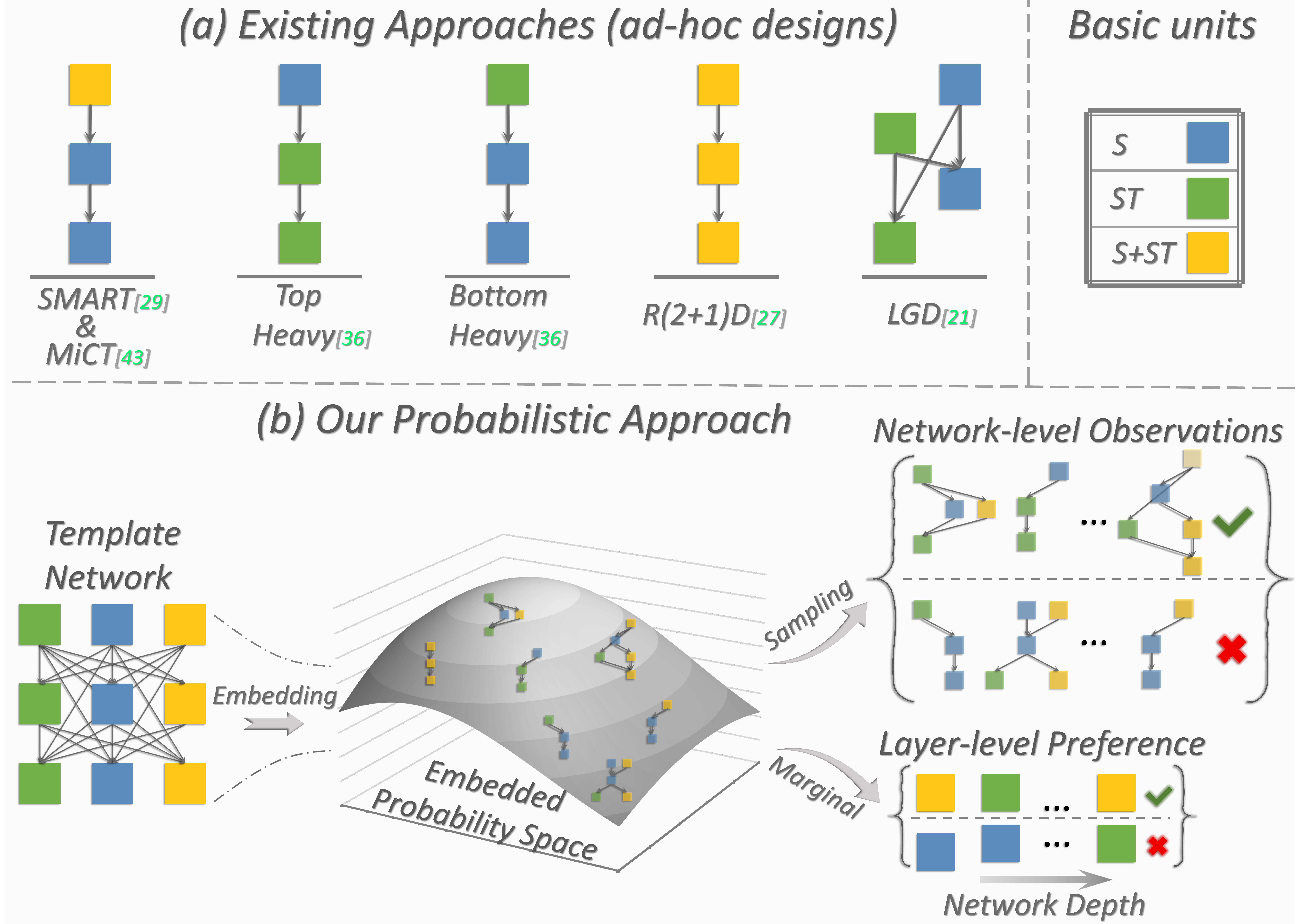}
    \caption{Spatiotemporal fusions in 3D CNNs. (a) Exemplified fusion methods reported in the literature, which are designed empirically and evaluated by training each corresponding network. (b) The proposed probabilistic approach. We propose to analyze the spatiotemporal fusion by finding a probability space where each individual fusion strategy is considered as a random event with a meaningful probability. We first introduce a template network based on basic fusion units to support a variety of fusion strategies. We then embed all possible fusion strategies into the probability space defined by the posteriori distribution over fusion strategy. As a result, various fusion strategies can be evaluated/analyzed without separate network training to obtain network-level observations and layer-level preference. Here $S$, $ST$ and $S+ST$ are basic fusion units instantiated by 2D, 3D, and a mix of 2D/3D convolutions, respectively.}
    \label{motivations}
\end{figure}

For numerous video applications, such as action recognition \cite{wang2016temporal, zhou2018mict, wu2019mutually}, video annotation \cite{zhou2017adaptive} and person re-identification \cite{zhang2018learning}, spatiotemporal fusion is an integral component. 
Taking action recognition as an example, the spatiotemporal fusion in deep networks can be roughly classified into two main categories: fusion/ensemble of two modalities (\textit{i.e}, spatial semantics in RGB and temporal dynamics in optical flow) in a two-stream architecture \cite{wang2016temporal, simonyan2014two} and fusion of spatial and temporal clues in single-stream 3D CNNs \cite{wang2018appearance, zhou2018mict}. In this paper, we focus on the latter. 
\blfootnote{$*$ This work was performed while Yizhou Zhou was an intern with Microsoft Research Asia. $\dagger$ Corresponding author.}

Conceptually, 3D CNNs are capable of learning spatiotemporal features responding to both appearance and movement in videos.  Recent research also shows that pure 3D CNNs can outperform 2D ones on large scale benchmarks \cite{hara2018can}. However, we still observe noticeable variations in accuracy by employing additional spatial or temporal feature learning explicitly in 3D CNNs. As shown at the top of Fig. \ref{motivations}, different spatiotemporal fusion strategies \cite{wang2018appearance, qiu2019learning, xie2018rethinking, tran2018closer, zhou2018mict} have been studied and recommended for action recognition. They explore spatial semantics and temporal dynamics in videos through the combinations of different types of basic convolution unit at each layer in 3D CNNs. Though with different conclusions, these works have one thing in common - they draw conclusions based on the performance of networks employing one or several fusion strategies designed empirically \cite{tran2018closer, xie2018rethinking, tran2017convnet}. Each fusion strategy is predefined, fixed, and evaluated in each individual network, leading to a network-level analysis of fusion strategies. Due to the proliferation of combinations and prohibitive computational costs, it is difficult for existing solutions to simulate a great number of fusion strategies for evaluation, nor can they support fine-grained and layer-level analysis. 

In this paper, we propose to analyze the spatiotemporal fusion in 3D CNNs from a different point of view, \textit{i.e.}, a probabilistic one. To be specific, we make the spatiotemporal fusion analysis an optimization problem, aiming to find a probability space 
where each individual fusion strategy is treated as a random event and assigned with a meaningful probability. The probability space will be constructed to meet the following requirements. First, the effectiveness of each spatiotemporal fusion strategy (event) can be \textit{easily} derived from the probability space, so that we can analyze all the fusion strategies based on the derived effectiveness rather than training each network defined by each fusion strategy. Second, from the probability which is closely correlated with the performance of each fusion strategy, it should be able to deduce the layer-level metrics of the fusion efficiencies, making it possible to perform layer-level, fine grained analysis of fusion strategies. 
Now, the question becomes how we build this probability space.

Recent research shows that optimizing a neural network with dropout (applied on every channel of kernel weights) is mathematically equivalent to the approximation to the posteriori distribution over the network weights \cite{gal2016dropout} and architectures \cite{zhou2019one}. It inspires us to construct the probability space via dropout in 3D CNNs. In our approach, we propose to first design a template network based on basic fusion units. We define the basic unit as different forms of spatiotemporal convolutions in 3D CNNs, \textit{e.g.}, spatial, spatiotemporal, and spatial+spatiotemporal convolutions, as illustrated in Fig. \ref{motivations}. 
The probability space can then be defined by the posteriori distribution on different sub-networks (fusion strategies) along with their associated kernel weights in the template network.
Note that in our fusion analysis, we need to approximate posteriori distribution on basic fusion units rather than on kernels as in \cite{gal2016dropout}. Therefore, based on the variational Dropout \cite{kingma2015variational} and DropPath \cite{larsson2016fractalnet}, we present a Variational DropPath (v-DropPath) by using a variational distribution which factorizes over the probability of the dropout operations that are applied on every basic fusion unit. Then the posterior distribution can be inferred by minimizing the Kullback-Leibler (KL) divergence between the variational distribution and the posteriori distribution, which proves to be equivalent to optimizing the template network with the v-DropPath. 
We will show that such a probability space fully satisfies the two requirements mentioned above in Section \ref{section_probability} and \ref{section_fusion}.  

Once we obtain such distribution, we acquire a variety of fusion strategies from the template network by executing v-DropPath \textit{w.r.t.} its optimized drop probability. Those fusion strategies can be directly evaluated without training. In addition, we also utilize the derived probability space to provide numerical measurements for layer-level spatiotemporal fusion preference. 

Experimental results show that our proposed probabilistic approach can produce very competitive fusion strategies to obtain state-of-the-art results on four widely used databases on action recognition. It also provides general and practical hints on the spatiotemporal fusion that can be applied to 3D networks with different backbones, such as ResNet\cite{he2016deep}, MobileNet\cite{sandler2018mobilenetv2}, ResNeXt\cite{xie2017aggregated} and DenseNet\cite{huang2017densely}, and achieve good performance.  

In summary, our work has four main contributions:
\begin{enumerate}
    \item We are the first to investigate the spatiotemporal fusion in 3D CNNs from a probabilistic view. Our proposed probabilistic approach enables a highly efficient and effective analysis on varieties of spatiotemporal fusion strategies. The layer-level fine-grained numerical analysis on spatiotemporal fusion also becomes possible.
    \item We propose the Variational DropPath to construct the desired probability space in an end-to-end fashion. 
    \item  New spatiotemporal fusion strategies are constructed based on the probability space and achieve the state-of-the-art performance on four well-known action recognition datasets. 
    \item We also show that the hints on spatiotemporal fusion obtained from the probability space are generic and suitable for benefiting different backbone networks.
\end{enumerate}

\section{Related Work}
Spatiotemporal fusion has been widely investigated in various tasks and frameworks \cite{qiu2019learning, liu2019dense, zhou2019context}.
In this paper, we choose one of its typical scenarios, \textit{i.e.}, action recognition, to discuss the related work. We further roughly group the spatiotemporal fusion methods for action recognition into two categories: fusion in two-stream (RGB and optical flow) CNNs and fusion in single 3D CNNs. Due to space limitations, here we review only the most related work - spatiotemporal fusion in single 3D CNNs. 

There exists a considerable body of literature on spatiotemporal fusion in 3D CNNs. Some of these works show that the efficiency of 3D CNNs can be improved by empirically decoupling the spatiotemporal feature learning in a specific way \cite{wang2018appearance, feichtenhofer2018slowfast, qiu2019learning, zhou2018mict, feichtenhofer2016spatiotemporal, zolfaghari2018eco, diba2018spatio, jiang2019stm}.  
For example, Wang et al. \cite{wang2018appearance} present the fusion method that utilizes 3D convolution with square-pooling to capture the appearance-independent relation and 2D convolution to capture the static appearance information. These two features are then concatenated and fed into a 1x1 convolution to form new spatiotemporal features. Results show that this fusion method can significantly improve the performance with model size and FLOPs similar to the original 3D architecture.
Feichtenhofer et al. \cite{feichtenhofer2018slowfast} also propose a fusion approach which combines the 3D and 2D CNNs. They use 2D convolution (with more channels) to capture rich spatial semantics from individual frames at lower frame rate, and factorized 3D convolution to extract motion information from frames at high temporal resolution which is fused by lateral connection to the 2D semantics.
Zhou et al. \cite{zhou2018mict} present a mixed 3D/2D convolutional tube, MiCT-block, which integrates 2D CNNs with 3D convolution via both concatenated and residual connections in 3D CNNs. It encourages each 3D convolution in 3D network to extract temporal residual information by adding its outputs to the spatial semantic features captured by 2D convolutions.  

Instead of presenting one specific fusion strategy, some other work investigates the spatiotemporal fusion in 3D CNNs by evaluating a group of pre-defined fusion methods 
\cite{tran2018closer, xie2018rethinking, tran2017convnet}. For instance, four fusion methods are constructed, trained and evaluated individually in \cite{xie2018rethinking} including bottom-heavy-I3D, top-heavy-I3D as shown in Fig.\ref{motivations}. More fusions such as mixed convolutions and reversed mixed convolutions are investigated in a similar way in \cite{tran2018closer, tran2017convnet}.
Although with meaningful observations, these methods can only analyze a limited number of fusion strategies, provide network-level hints, and suffer from huge computational costs.

In contrast to all the above presented methods, in this paper, we propose to construct a probabilistic space that encodes all possible spatiotemporal fusion strategies under a predefined network topology. It not only provides a much more efficient way to analyze a variety of fusion strategies without training them individually, but also facilitates the fine-grained numerical analysis on the spatiotemporal fusion in 3D CNNs.

\section{Spatiotemporal Fusion in Probability Space}\label{section_symbo}

\begin{figure}
    \centering
    \includegraphics[width=\linewidth]{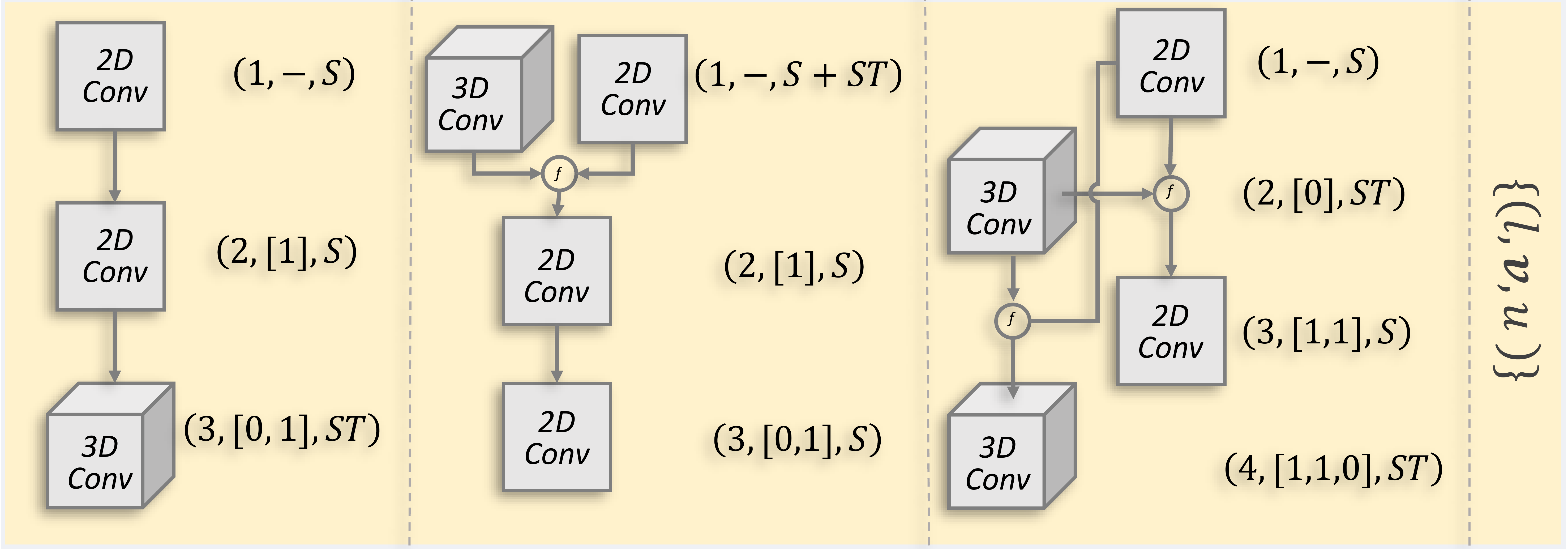}
    \caption{Exemplified triplet representations $ \{(l, \mathbf{v}, u) \}$ of three spatiotemporal fusion strategies reported in literature.}
    \label{symbolization}
\end{figure}

We observe that a fusion strategy in an $L$-layer 3D CNN can be expressed with a set of triplets $ \{(l, \mathbf{v}, u) \}_L$, where $l$ ($1 \leq l \leq L $) is the layer index, $\textbf{v}$ is a binary vector of length $l-1$ denoting the features from which layer/layers will be used, and $u$ ($u\in U$) denotes the basic fusion units employed in the current layer. Here $U$ is defined by a set of basic fusion units. For example, $U$ can be the combination of three modes, Spatial (S), temporal (T), and spatiotemporal (ST), \textit{i.e.}, $U = \{S,T,ST,S+T,S+ST,T+ST,S+T+ST\}$. 
As concrete examples, existing fusion strategies can be well represented by the triplets, \textit{e.g.}, top-heavy structure \cite{xie2018rethinking}, SMART-block\cite{wang2018appearance}/MiCT-block \cite{zhou2018mict} and global diffusion structure \cite{qiu2019learning}, as shown in Fig. \ref{symbolization}, respectively.

\subsection{The Probability Space}\label{section_probability}
As discussed in the introduction, we construct the probability space with the posteriori distribution over different fusion strategies along with their associated kernel weights. In the probability space, $\mathcal{M}=\{(l, \mathbf{v}, u)\}_{L}$ should be a random event. We also define $W_{\mathcal{M}}$ to be the kernel weight of the corresponding strategy $\mathcal{M}$, which is also a random event in such space. Therefore, we give the full definition of the probability space denoted with $(\Omega, \mathcal{B}, \mathcal{F})$, where
\begin{itemize}
    \item Sample space $\Omega = \{(\mathcal{M}, W_{\mathcal{M}})\}$, which is the set of all possible outcomes from the probability space.
    \item A set of events $\mathcal{B} = \{(\mathcal{M}, W_{\mathcal{M}})\}$, where each event is equivalent to one outcome in our case.
    \item Probability measure function $\mathcal{F}$. We use the posteriori distribution to assign probabilities to the events as
    \begin{equation}
    \mathcal{F} := \mathcal{P}(\mathcal{M}, W_{\mathcal{M}} \mid \mathbb{D}),
    \label{Eq_Posteriori}
    \end{equation}
    where $\mathbb{D}=\{X, Y\}$ indicates the data samples $X$ and ground-truth label $Y$ used for training.
\end{itemize}

In this probability space, various fusion strategies and their associated kernel weights are sampled as pairs and we can make direct evaluation without training. The overall performance of one strategy can be obtained only at the cost of network testing. Therefore, the first requirement for the probability space is satisfied. Now, The core of embedding spatiotemporal fusion strategies into such probability space is to derive the measure function defined in Eq. \ref{Eq_Posteriori}.

\subsection{Embedding via Variational DropPath}\label{section_embedding}
It is hard to obtain the posteriori distribution in Eq. (\ref{Eq_Posteriori}), as usual. In our approach, we present a variational Bayesian method to approximate it. We first build a template network based on the basic fusion units that will be studied in the spatiotemporal fusion. For instance, we can design a densely connected 3D CNN with $ U=\{ S, ST, S+ST \}$, as shown in Fig. \ref{motivations}.
We then incorporate a variational distribution that factorizes over every basic unit in the template network, which are re-parameterized with kernel weight multiplying a dropout rate. 
We further propose the v-DropPath inspired by \cite{kingma2015variational,gal2016dropout, zhou2019one} that enables us to minimize the KL distance between the variational distribution and the posteriori distribution via training the template network. More details will be presented below. 

By incorporating the template network, the posterior distribution in Eq. (\ref{Eq_Posteriori}) can be converted to
\begin{equation}
     \mathcal{P}(\mathcal{M}, W_{\mathcal{M}} \mid \mathbb{D}) \xrightarrow[]{} \mathcal{P}(\mathcal{\widehat{M}} \circ W_T \mid \mathbb{D}),
\end{equation}
where $\circ$ is the Hadamard product (with broadcasting), $\mathcal{\widehat{M}} \in (0,1)^{L \times L \times 3}$ is a binary random matrix
and $\mathcal{\widehat{M}}(l, i, u)=1/0$ denotes that the feature from the layer $i$ and the fusion unit $u$ is enabled/disabled at layer $l$ in the template network, respectively.
$W_T \in \mathbb{R}^{L \times L \times 3 \times V}$ denotes the random weight matrix of the template network, where we use $V$ to denote kernel shape for simplicity. This conversion actually integrates the kernel weights into fusion strategies. Since we can fully recover the $\mathcal{M}$ from the embedded version $\mathcal{\widehat{M}} \circ W_T$ (it is because the kernel is defined in real number field, the probability of being zero for every element can be ignored), the first requirement is still satisfied.

We then approximate the posteriori distribution by minimizing the KL divergence
\begin{equation}\label{KLterm}
    KL(\mathcal{Q}(\mathcal{\widehat{M}} \circ W_T) \mid \mid \mathcal{P}(\mathcal{\widehat{M}} \circ W_T \mid \mathbb{D})),
\end{equation}
where $\mathcal{Q}(\cdot)$ denotes a variational distribution.
Instead of factorizing the variational distribution over convolution channels as in \cite{gal2016dropout}, we factorize $\mathcal{Q}(\mathcal{\widehat{M}} \circ W_T)$ over fusion units in each layer as 
\begin{equation}\label{factorization}   
    \prod_{l, i, u} q(\mathcal{\widehat{M}}(l, i, u) \cdot W_T(l, i, u, :)).
\end{equation}
By re-parameterising the $q(\mathcal{\widehat{M}}(l, i, u) \cdot W_T(l, i, s, :))$ with $\epsilon_{l,i,u} \cdot w_{l,i,u}$, where $\epsilon_{l,i,u} \sim Bernoulli(p_{l,i,u})$ and $w_{l,i,u}$ is the deterministic weight matrix associated with the random weight matrix $ W_T(l, i, u, :)$, minimizing Eq. \ref{KLterm} is approximately equivalent to minimizing 
\begin{equation}\label{objective}
    \begin{split}
        - \frac{1}{N} &\log \mathcal{P}(Y \mid X, w \cdot \epsilon) + \frac{1}{N}\sum_{l,i,u} p_{l,i,u} \log p_{l,i,u} \\ &+\sum_{l,i,u}\frac{(k_{l,i,u})^2(1-p_{l,i,u})}{2N} \Arrowvert w_{l,i,u} \Arrowvert^2,
    \end{split}
\end{equation}
where $k_{l,i,u}$ is a pre-defined length-scale prior and $N$ is the number of training samples. The gradients w.r.t. the Bernoulli parameters $p$ are computed through Gumbel-Softmax \cite{jang2016categorical}.  For step-by-step proofs of Eq. \ref{objective}, please refer to our supplementary material. 

Eq. \ref{objective} reveals that approximating the posteriori distribution can be achieved by training the template 3D network where each spatial or temporal convolutions is masked by a logit $\epsilon$ subject to Bernoulli distribution with probability $p$. It is exactly the drop-path proposed in \cite{larsson2016fractalnet}. But here both the network weight and the drop rate need to be optimized. We adopt Gumbel-Softmax for the indifferentiable Bernoulli distribution to enable a gradient-based solution. Please find more details in supplementary material.

\subsection{Spatiotemporal Fusion}\label{section_fusion}
Once the probability space defined by the posteriori distribution is obtained, we can investigate the spatiotemporal fusion very efficiently at both the network and layer levels.

\textbf{Network-level}. Conventionally, the network-level fusion strategies are explored by training and evaluating each individual network defined by one fusion strategy. 
In our scheme, we successfully eliminate the individual training and evaluation by using the embedded probability space. We study the fusion strategies by directly sampling a group of strategy and kernel weight pairs $\{(\mathcal{M}, W_{\mathcal{M}})^t \mid t=1,2,...\}$ with
\begin{equation}
    \mathcal{M}, W_{\mathcal{M}} \sim  \mathcal{P}(\mathcal{\widehat{M}} \circ W_T \mid D_{tr}) \approx \mathcal{Q}(\mathcal{\widehat{M}} \circ W_T).
\end{equation}
It is doable since each $(\mathcal{M}, W_{\mathcal{M}})^t$ can be fully recovered from the embedded version $\mathcal{\widehat{M}} \circ W_T$. The above sample process is equivalent to randomly choosing $\epsilon_{l, i, u}$ based on the Bernoulli distribution with the optimized $p_{l, i, u}$ as defined in Eq. \ref{objective}, which is further equivalent to randomly dropping some paths in the template network. The effectiveness of each fusion strategy can then be easily derived from the test performance on a validation dataset. Because the sampling and evaluation are light-weight, our approach can greatly expand both the number and form of fusion strategies for analysis. 

\textbf{Layer-level}. The network-level analysis shows the overall effectiveness of different spatiotemporal fusion strategies, but rarely reveals the importance of the fusion strategies at each layer.
Interestingly, numerical metrics for such fine-grained, layer-level information are also achievable in our approach. Recall that we factorize the variational distribution in Eq. \ref{factorization} over different fusion strategies using the reparametrisation trick \cite{kingma2015variational}. We thus can deduce the marginal probability of fusion unit at each layer as
\begin{equation}\label{marginal}
    \mathcal{P}(\mathcal{\widehat{M}}(l, i, u)=1 \mid \mathbb{D}) = 1 - \sqrt{p_{l,i,u}}.
\end{equation}
Please refer to supplementary material for detailed derivation. Eq. \ref{marginal} suggests that the marginal distribution of a spatiotemporal fusion strategy can be retrieved from the optimized dropout probability. It indicates the probability of using a fusion unit among all the possible networks that can interpret the given dataset well and satisfy prior constrains (sparsity in our case).  We propose using this number as the indicator of the layer-level spatiotemporal preference. Therefore, the second requirement on the probability space is met, too.

\section{Experiments}
In this section, we will verify the effectiveness of our probabilistic approach from three aspects. Four action recognition databases are used in the experiments. After the description of experimental setups, we will first show the performance of the fusion strategies obtained by our approach in comparison with those of state-of-the-arts. Then several main observations are provided based on the analysis of different fusion strategies generated from our probability space. At last, we verify the robustness of the obtained spatiotemporal fusion strategies on different backbone networks. 

\begin{table*}
\vspace{-4mm}
\centering
\small
\caption{Performance evaluation on Something-Something V1. Im./K.400 denote ImageNet/Kinetics400 pre-training.}
\vspace{-2mm}
\begin{tabular}{l l c c c c c c c}
\hline
\hline
Method & Backbone & Extra Mod. & Pretrain & \#F & FLOPs & \#Param. & Top-1 & Top-5 \\
\hline
\hline

TSN\cite{wang2016temporal}  & BNInception & - & Im. & 8 & 16G & 10.7M & 19.5\% & - \\
TSN\cite{lin2019tsm}  & ResNet50 & - & Im. & 8 & 33G & 24.3M & 19.7\% & 46.6\% \\
TRN-Multiscale\cite{zhou2018temporal}  & BNInception & - & Im. & 8 & 16G & 18.3M & 34.4\% & - \\
TRN-Multiscale\cite{lin2019tsm}  & ResNet50 & - & Im. & 8 & 33G & 31.8M & 38.9\% & 68.1\% \\
Two-stream TRN\cite{zhou2018temporal}  & BNInception & - & Im. & 16 & - & 36.6M & 42.0\% & - \\
TSM\cite{lin2019tsm}  & ResNet50 & - & Im. & 16 & 65G & 24.3M & 47.2\% & 77.1\% \\
\hline
TrajectoryNet\cite{zhao2018trajectory} & 3D Res.18 & Y & Im.+K.400 & - & - & x & 47.8\% & - \\
STM\cite{jiang2019stm} & 3D Res.50 & Y & Im. & 16 & 66.5G & 24.0M & 49.8\% & - \\
Non-local I3D\cite{wang2018videos} & 3D Res.50 & Y & Im. & 64 & 336G & 35.3M & 44.4\% & 76.0\% \\
Non-local I3D + GCN\cite{wang2018videos} & 3D Res.50+GCN & Y & Im. & 64 & 606G & 62.2M & 46.1\% & 76.8\% \\
S3D-G\cite{xie2018rethinking} & 3D BNincept.+gate & Y & Im. & 64 & 71G & 11.6M & 48.2\% & 78.7\% \\
\hline
I3D\cite{wang2018videos} & 3D Res.50 & N & Im. & 64 & 306G & 28.0M & 41.6\% & 72.2\% \\
I3D\cite{xie2018rethinking} & 3D BNIncept. & N & Im. & 64 & 108G & 12.0M & 45.8\% & 76.5\% \\
S3D\cite{xie2018rethinking} & 3D BNIncept. & N & Im. & 64 & 66G & 8.77M & 47.3\% & 78.1\% \\
ECO\cite{zolfaghari2018eco} & BNIncept.+3DRes.18 & N & Im.+K.400 & 8 & 32G & 47.5M & 39.6\% & -\\
ECO\cite{zolfaghari2018eco} & BNIncept.+3DRes.18 & N & Im.+K.400 & 16 & 64G & 47.5M & 41.4\% & - \\
ECO Lite\cite{zolfaghari2018eco} & BNIncept.+3DRes.18 & N & Im.+K.400 & 92 & 267G & 150M & 46.4\% & - \\
\hline

Ours & 3D DenseNet121 & N & Im. & 16 & 31G & 21.4M & \textbf{50.2\%} & \textbf{78.9\%} \\

\hline
\hline
\end{tabular}
\label{somethingv1_stateart}
\end{table*}

\subsection{Experimental Setups}
\begin{figure}
    \centering
    \includegraphics[width=\linewidth]{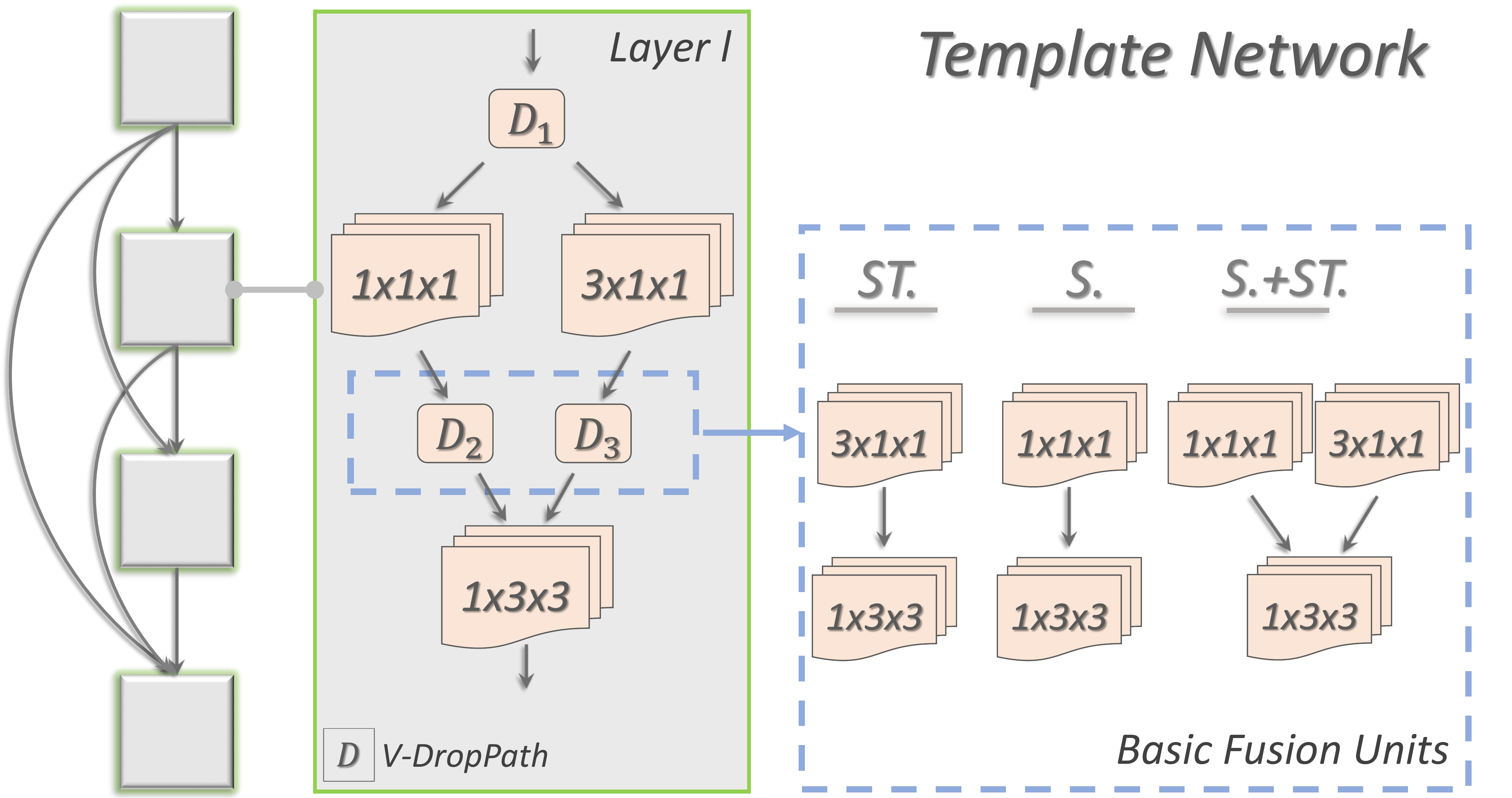}
    \caption{The densely connected template network used in our experiments. In each layer, there are three DropPath (D) operations. The combination of $D_2$ and $D_3$ deduces the three basic fusion units \{$S,ST,$ and $S+ST$ \}. The operations on $D_1$ and $D_2$/$D_3$ correspond to the index $i$ and $u$ in $\epsilon_{l,i,u}$, respectively.}
    \label{template}
\end{figure}

\textbf{Template Network.} Fig. \ref{template} sketches the basic structure of the template network designed for our approach. The template network is a densely connect one that comprises of mixed 2D and 3D convolutions. Here we choose $U=\{ S,ST,S+ST \}$ so that the fusion units explored in our approach are conceptually included in most of other fusion methods for fair comparison.
We also factorize each 3D convolution with a 1D convolution and a 2D convolution, and use element-wise summation to fuse the 2D and 3D convolutions for simplicity. Besides, we add several transition blocks to reduce the dimension of features and the total number of layers is set to be 121 as in \cite{huang2017densely}. We put more details of the template network in the supplementary material. In practice, we share the variational probability on the variables $i$ defined in Section. \ref{section_symbo} for computational efficiency.

\textbf{Datasets.} We apply the proposed scheme on four well-known action recognition datasets, \textit{i.e.}, Something-Something(V1\&V2)\cite{goyal2017something}, Kinetics400\cite{kay2017kinetics} and UCF101\cite{soomro2012ucf101}. Something V1/V2 consist of around 86k/169k videos for training and 12k/25k videos for validation, respectively. Video clips in these two datasets are first-person videos with 174 categories that focus more on temporal modelling. Kinetics400 is a large-scale action recognition database which provides around 240k training samples and 20k validation samples from 400 classes. UCF101 has around 9k and 3.7k videos for training and validation. They are categorized into 101 classes. Both the Kinetics400 and the UCF101 contain complex scene and object content in video clips and have large temporal redundancy.

\textbf{Training.}
As mentioned before, we approximate the posteriori distribution of different fusion strategies by training the template network with v-DropPath. We initialize the drop rate of each convolution operation as $0.1$. We train the template network with 90 epochs for Something-Something(V1\&V2)/UCF101 and 110 epochs for Kinetics400, respectively. The batch size is 64 for Kinetics and 32 for the others. The initial learning rates are 0.005 (Something\&UCF) and 0.01 (Kinetics) and we decay them by multiplying 0.1 at 40th, 60th, 80th epochs for Something/UCF and 40th, 80th epochs for Kinetics. The video frames are all resized to 256 (short edge) and randomly cropped to 224x224. The length-scale prior $k$ in Eq. \ref{objective} is determined by grid search, where $k=250$ for SomethingV1, $k=10$ for Kinetics400 and $k=50$ for the rest. In practice, warmup is used before training the template network with v-DropPath, \textit{i.e.}, removing all the v-DropPath operations and training the template network from scratch for 50 epochs. All experiments are conducted with distributed settings and synchronized Batch Normalization \cite{ioffe2015batch} on multiple (8-32) v100 GPUs with 32G memory.

\textbf{Sampling and Inference.}
We derive various spatiotemporal fusion strategies from the probability space through sampling different combinations of spatiotemporal convolutions w.r.t. the drop probability of v-DropPath. The sampled strategies are directly evaluated on validation dataset. 

During the inference of each spatiotemporal fusion strategy, we resize the short edge of the input video frames to 256 and make center crop to get a 256x256 region. We uniformly sample multiple clips in a video and average the prediction scores to obtain video level predictions. The number of clips varies from dataset to dataset and will be discussed along with the results.

\subsection{Ablation Study}\label{section_abliation}
In order to demonstrate the effectiveness of our probability space, for each dataset, we sample 100 fusion strategies from the constructed space and choose the best one according to the performance on the held-out validation dataset. We denote the best strategy as `Optimized'(Opt). We then compare it with its counter-part strategies `S',`ST', and `S+ST' in Fig. \ref{ablation}, which are designed with one fixed corresponding basic fusion unit, $S, ST$, or $S+ST$, at all layers, respectively. It can be observed that our probability space can generate better strategies on all the dataset. Our `Opt' method even outperforms its counter-part `ST+S' which has more parameters and higher FLOPs.

\begin{table}[]
\small{}
    \centering
    \caption{Ablation studies on the selected spatiotemporal fusion strategies from our probability space.}
    \vspace{-2mm}
    \begin{tabular}{|l||*{4}{c|}}\hline
            \backslashbox{Dataset}{Strategy}
            & S & ST & S+ST & Opt \\ \hline\hline
            SomethingV1 & 41.8\% & 47.5\% & 46.5\% & \textbf{50.2}\% \\ \hline
            SomethingV2 & 55.1\% & 60.5\% & 59.5\% & \textbf{62.4}\% \\  \hline
            UCF101 & 83.6\% & 83.1\% & 84.2\% & \textbf{84.2}\% \\  \hline
            Kinetics400 & 67.8\% & 68.3\% & 69.7\% & \textbf{71.7}\% \\ \hline
    \end{tabular}
    \label{ablation}
\end{table}

\subsection{Comparisons with the State-of-the-arts}
Our proposed method analyzes the spatiotemporal fusion strategies from the perspective of the probability. It not only enables an advance analysis approach, but also achieves high-performance spatiotemporal fusion strategies. In this section, we compare the strategies drawn from the probability space with state-of-the-art fusion methods on four action recognition datasets. Our approach has very competitive performance, \textit{i.e.}, performing the best among all the schemes on three of these datasets and obtaining the second best on UCF101, even though some of the compared results are achieved with better backbones and/or with extra modules such as non-local, motion encoder, or gated functions.

\textbf{Something-Something V1\&V2.} 
Table. \ref{somethingv1_stateart} exhibits the performance of different spatiotemporal fusion methods on Something V1 dataset. It shows that our approach leads to the fusion strategy that outperforms all the other schemes including so far the most advanced 3D network S3D by a large margin with 50\% fewer FLOPs and frames. Surprisingly, it performs even better than those methods with carefully designed functional modules, \textit{e.g.}, STM employs a channel-wise motion module to explicitly encode motion information, and Non-local I3D + GCN explicitly incorporates the object semantics with graphs. Similar results can be observed on the recently released dataset Something V2. As shown in Table. \ref{somethingv2_stateart}, our fusion strategies significantly outperform the conventional I3D solutions and its bottom-heavy and top-heavy counterparts which incorporates 3D convolutions in bottom layers and top layers, respectively. We employ ImageNet pre-training for both datasets and our fusion strategy can achieve higher accuracy than those pre-trained on the large-scale dataset Kinetics such as ECO.

\begin{table}
\vspace{-2mm}
\centering
\small
\caption{Performance comparison with state-of-the-art results on Something-Something V2.}
\vspace{-2mm}
\begin{tabular}{l c c}
\hline
\hline
Method & Val. Top-1 & Val. Top-5 \\
\hline
\hline

TSN\cite{lin2019tsm} & 30.0\% & 60.5\% \\
MultiScale TRN\cite{zhou2018temporal} & 48.8\% & 77.6\% \\
Two-stream TRN\cite{zhou2018temporal} & 55.5\% & 83.1\% \\
\hline
TSM(ImageNet+ Kinetics400)\cite{lin2019tsm} & 59.1\% & 85.6\% \\
TSM dual attention\cite{xiao2019reasoning} & 55.5\% & 82.0\% \\
I3D-ResNet50\cite{xiao2019reasoning} & 43.8\% & 73.2\% \\
2D-3D-CNN w/ LSTM \cite{mahdisoltani2018fine} & 51.6\% & - \\
Ours (ImageNet) & \textbf{62.9}\% & \textbf{88.0}\% \\

\hline
\hline
\end{tabular}
\label{somethingv2_stateart}
\end{table}

\textbf{Kinetics400.} 
Accuracy achieved by different fusion methods on Kinetics400 are reported in Table \ref{kinetics_stateart}. 
In order to make apple-to-apple comparisons, all methods are trained from scratch. It can be observed that our configuration of spatiotemporal fusion outperforms the second best R(2+1)D on Top1 accuracy with 97\% fewer FLOPs , where R(2+1)D is a 3D network that uses ResNet34 as backbone. Compared with R(2+1)D, we actually utilize more spatial convolutions in the shallow layers as can be viewed in Fig. \ref{kinetics_stateart}.

\begin{table}
\small
\centering
\caption{Performance comparison with the state-of-the-art results of different spatiotemporal fusions in 3D architectures on Kinetics400 trained from the scratch.}
\vspace{-2mm}
  \begin{tabular}{l|l|l|l|l}
    \hline
    \hline
    Method &Backbone & FLOPs & Top1 & Top5  \\
    \hline
    \hline
    STC\cite{diba2018spatio} & R.Xt101 & N/A $\times$ N/A & 68.7\% & 88.5\% \\
    ARTNet\cite{wang2018appearance} & ResNet18 & 23.5G $\times$ 250 & 69.2\% & 88.3\% \\
    R(2+1)D\cite{tran2018closer} & ResNet34 & 152G $\times$ 115 & 72.0\% & 90.0\% \\
    S3D*\cite{xie2018rethinking} & BNIncept. & 66.4G $\times$ 250 & 69.4\% & 89.1\% \\
    I3D\cite{carreira2017quo} & BNIncept. & 216G $\times$ N/A & 68.4\% & 88.0\% \\
    ECO\cite{zolfaghari2018eco} & custom & N/A $\times$ N/A & 70.0\% & 89.4\% \\
    3DR.Xt\cite{hara2018can} & R.Xt101 & N/A $\times$ N/A & 65.1\% & 85.7\% \\
    Disentan.\cite{zhao2018recognize} & BNIncept. & N/A $\times$ N/A & 71.5\% & 89.9\% \\
    StNet \cite{he2019stnet} & ResNet101 & 311G x 1 & 71.4\% & - \\
    \hline
    
    Ours  & Dense.121 & 254G $\times$ 2 & \textbf{72.5}\% & \textbf{90.3}\% \\
    
    \hline
    \hline
    
  \end{tabular}
  \label{kinetics_stateart}
\vspace{-2mm}
\end{table}

\textbf{UCF101.} 
Since UCF101 has only 9k training videos, we make evaluations with the ImageNet pre-training and Kinetics400 pre-training, respectively. When incorporating ImageNet pre-training only, our fusion strategy produces the most advanced results, which has $1.5\%$ higher accuracy than the I3D that performs pure spatiotemporal fusions. When using Kinetics400 as pre-training dataset, the overall performance is still state-of-the-art. Please note that we do not employ any extra functional module here, so the performance is slightly worse ($0.3\%$) than the most advanced 3D networks S3D-G that incorporates attention mechanism.

\begin{table}
\small
\caption{Performance comparison with the state-of-the-art results on UCF101. Im., S.1M and K.400 denote ImageNet, Sport1M and Kinetics400, respectively. Our methods with ResNeXt50 and Inception backbones are designed according to the hints we observe from the probability space. Please refer to Section \ref{observations} and \ref{sec_generalization} for details.}
\vspace{-2mm}
\centering
  \begin{tabular}{l|c|l|c}
    \hline
    \hline
    Method & Pre. & Backbone & Top-1 \\
    \hline
    \hline
    TDD\cite{wang2015action} & Im. & VGG-M & 82.8\% \\
    C3D\cite{tran2015learning} & Im. & 3DVGG11 & 44.0\%\\
    LTC\cite{varol2017long} & Im. & 3DVGG11 & 59.9\% \\
    ST-ResNet\cite{feichtenhofer2016spatiotemporal} & Im. & 3DRes.50 & 82.3\% \\
    I3D\cite{carreira2017quo} & Im. & 3DIncept. & 84.5\% \\
    Ours & Im. & 3DDenseNet121 & \textbf{85.0\%} \\
    Ours & Im. & 3DRexNeXt50 & \textbf{86.0\%} \\
    \hline
    
    Res3D\cite{tran2017convnet} & S.1M & 3DRes.18 & 85.8\% \\
    P3D\cite{qiu2017learning} & S.1M & 3DRes.199 & 88.6\% \\
    MiCT\cite{zhou2018mict} & S.1M & 3DIncept. & \textbf{88.9\%} \\
    \hline
    
    Res3D\cite{tran2017convnet} & K.400 & 3DRes.18 & 89.8\% \\
    TSN\cite{wang2016temporal} & K.400 & Incept.V3 & 93.2\% \\
    I3D\cite{carreira2017quo} & K.400 & 3DIncept. & 95.6\% \\
    ARTNet\cite{wang2018appearance} & K.400 & 3DRes.18 & 94.3\% \\
    R(2+1)D\cite{tran2018closer} & K.400 & 3DRes.34 & \textbf{96.8\%} \\
    S3D-G\cite{xie2018rethinking} & K.400 & 3DIncept. & \textbf{96.8\%} \\
    3DResNeXt101\cite{hara2018can} & K.400 & - & 94.5\% \\
    STM\cite{jiang2019stm} & K.400 & 3D Res.50 & 96.2\% \\
    STC\cite{diba2018spatio} & K.400 & 3DResNext101 & 96.5\% \\
    Ours & K.400 & 3DDenseNet121 & 94.5\% \\
    Ours & K.400 & 3DIncept. & 96.5\% \\
    
    \hline
    \hline
    
  \end{tabular}
  \label{ucf101_stateart}
\vspace{-2mm}
\end{table}

\subsection{Observations}\label{observations}
\begin{figure*}
\vspace{-6mm}
    \centering
    \includegraphics[width=0.95\linewidth]{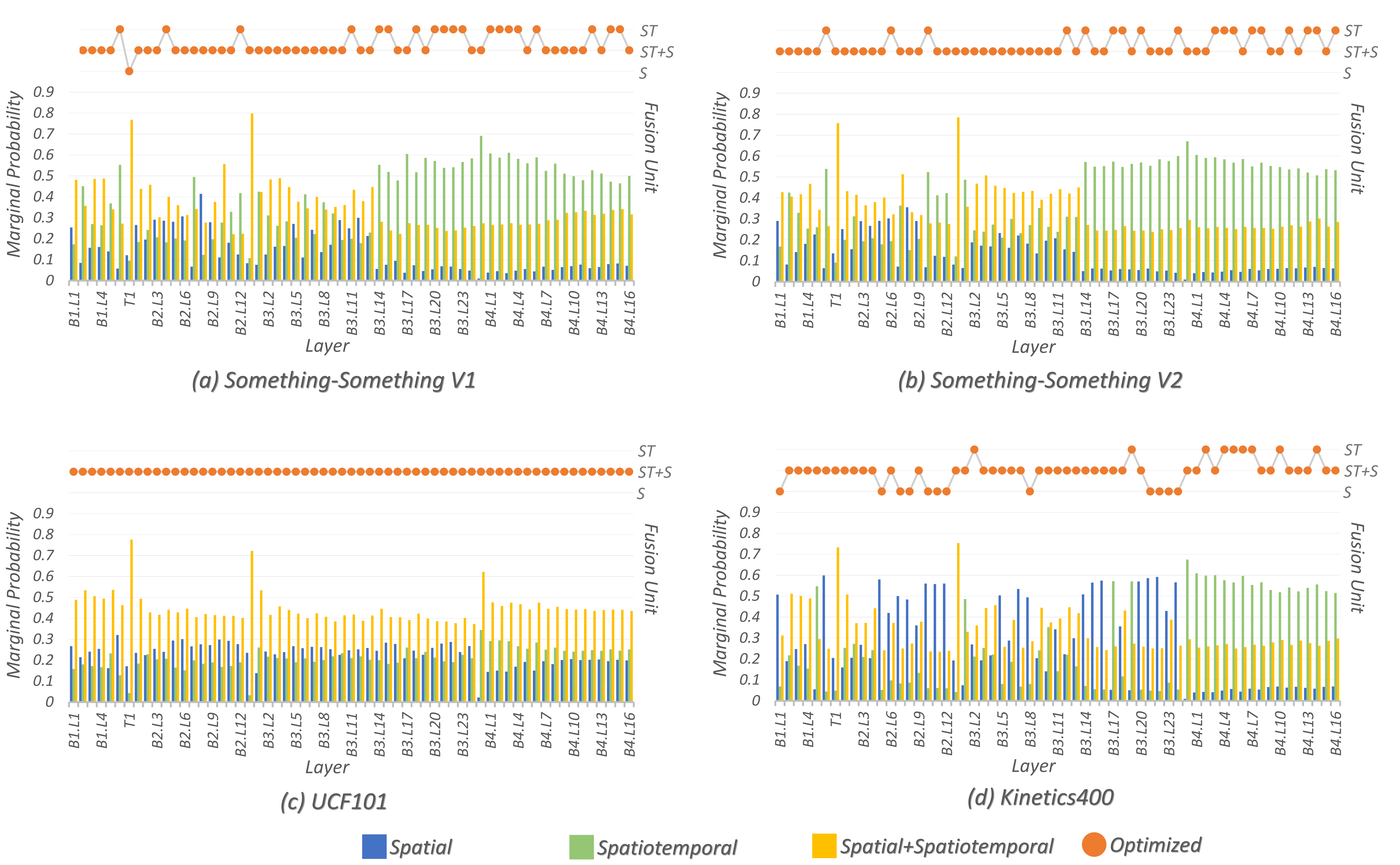}
    \caption{Visualization of our spatiotemporal fusion strategies and marginal probabilities of layer level fusion units. On the top of each sub-figure, we show the fusion strategy derived from the probability space that has the highest accuracy by the dotted line in orange. Three units, $S$, $ST$ and $S+ST$, are involved as shown on the right side of each sub-figure.
    The amplitude of bars in blue, green and yellow indicates the marginal probability of using the basic units $S$, $ST$ and $S+ST$ in each layer, respectively. The x-axis indexes layers, where B denotes dense blocks and L is the layer index in the block.}
    \label{fig:visualization}
\vspace{-2mm}
\end{figure*}

We visualize the strategies derived from the probability space that have the highest accuracy on the test datasets in Fig. \ref{fig:visualization}. We also illustrate the marginal probability of using different basic units in each layer based on Eq. \ref{marginal}. The amplitude of bars in blue, green and yellow indicates the marginal probability of using the units $S$, $ST$ and $S+ST$ in each layer, respectively. The dotted-line in orange shows the selected layer-level basic fusion units that produce the best accuracy. From these figures, we observe that

\textbf{Observation \RNum{1}}. As indicated by the colored bars, the unit $S+ST$ has higher marginal probability in the lower-level feature learning compared with the other two units. The dotted line in orange also shows a similar trend. The $S+ST$ unit has the highest percentage of total usage in all the fusion units, especially in the lower layers. It suggests that a proper spatiotemporal fusion strategy can be designed based on $S+ST$ units, particularly in lower layers.

\textbf{Observation \RNum{2}.} More $ST$ units are preferred in higher layers as there is a higher marginal probability on the $ST$ unit in the higher-level feature learning (except on UCF101 which will be discussed below). 

\textbf{Observation \RNum{3}.} Additional $S$ units could be beneficial when scene semantics are complex. For instance, Kinetics400/UCF101 contain videos in the wild with 400/101 different categories, respectively. The scene content is more complex than that in the first-person videos in Something-Something. By comparing Fig. \ref{fig:visualization} (c) and (d) with the others, it shows that more $S$ or $S+ST$ units are selected.

\subsection{Generalization}\label{sec_generalization}

\begin{table}[]
    \centering
    \centering
        \small
        \caption{Generalization of the observations. The fusion strategies `Opt' for each backbone are straightforwardly designed based on the observations.}
        \vspace{-2mm}
        \begin{tabular}{|l||*{4}{c|}}\hline
            \backslashbox{Net.}{Strategy}
            & S & ST  & S+ST & Opt \\ \hline\hline
            3D ResNet50 & 33.8\% & 40.1\% & 38.9\% & \textbf{41.2}\% \\  \hline
            3D ResNeXt50 & 35.2\% & 42.1\% & 40.7\% & \textbf{43.6}\%  \\  \hline
            3D ResNeXt101 & 36.6\% & 42.7\% & 42.3\% & \textbf{44.0}\%  \\ \hline
        \end{tabular}
    \label{gneralization}
    \vspace{-3mm}
\end{table}

We further discuss the generalization of our observations as well as the selected fusion strategies. We extend our fusion strategies to three backbone networks including ResNet50\cite{he2016deep}, and ResNeXt50/ResNeXt101\cite{xie2017aggregated}. They differ from each other in terms of topology, parameter size and FLOPs. We report clip-level accuracy on Something V1 for quick comparison. Please find more results and discussions on other backbone networks in the supplementary material.

We employ four different fusion strategies `Opt', `S+ST', `S' and `ST' as defined in Section \ref{section_abliation} for comparison. Note that here the fusion strategy denoted by ‘Opt' is not optimized using our probabilities approach but straightforwardly designed based on our observations. Specifically, we construct the fusion strategy `Opt' according to Fig. \ref{fig:visualization} (a) and (b), which uses $S+ST$ unit in both the first half and the last three layers, and $ST$ unit in the remaining layers. As shown in Table. \ref{gneralization}, the fusion method `Opt' performs the best among all the evaluated fusion strategies.

\section{Conclusion and Discussion}
In this paper, we convert the problem of analyzing spatiotemporal fusion in 3D CNNs into an optimization problem which aims to embed all possible fusion strategies into the probability space defined by the posteriori distribution on each fusion strategy along with its associated kernel weights. Such probability space enables us to investigate spatiotemporal fusion from a probabilistic view, where various fusion strategies are evaluated and analyzed without the needs of individual network training. The numerical measurements on layer-level fusion preference are available. By further proposing the Variational DropPath, the optimization problem can be efficiently solved via training a template network. Experimental results on four action recognition databases demonstrate the effectiveness of our approach. We also observe several useful hints with our probabilistic approach which can be extended to design high performance fusion strategies on different backbones.

\section*{Acknowledgement}
This work was supported by the National Key R\&D Program of China under Grant 2017YFB1300201, the National Natural Science Foundation of China (NSFC) under Grants 61622211, U19B2038 and 61620106009 as well as the Fundamental Research Funds for the Central Universities under Grant WK2100100030.

{\small
\bibliographystyle{ieee_fullname}
\bibliography{egbib}
}



\section*{Supplementary material}

# Supplementary Materials

The supplementary materials give the following information, which is not included in the submitted paper due to page limitation.

- Detailed proofs on Section 3.2 and 3.3 in the paper
- More details on experimental setup.
- More results on the Section 4.5 (Generalization).
- Discussions on Neural Architecture Search (NAS).

## 1. Detailed Proof

### 1.1. Equation 5 in the paper

By incorporating the template network, the probability measure function in Eq. (1) in the paper can be converted to

$$\mathcal{P}(\mathcal{M}, W_\mathcal{M} \mid \mathbb{D}) \rightarrow \mathcal{P}(\widehat{\mathcal{M}} \circ W_T \mid \mathbb{D}), \quad \text{(S1)}$$

where $\circ$ is the Hadamard product[1], $\widehat{\mathcal{M}} \in (0,1)^{L \times L \times 3}$ is a binary random matrix and $\widehat{\mathcal{M}}(l, i, u) = 1/0$ denotes that the feature from the layer $i$ and the fusion unit $u$ is enabled/disabled at layer $l$ in the template network, respectively. $W_T \in \mathbb{R}^{L \times L \times 3 \times V}$ denotes the random weight matrix of the template network, where we use $V$ to denote kernel shape for simplicity. This conversion actually integrates the kernel weights into fusion strategies. Since we can fully recover the $\mathcal{M}$ from the embedded version $\widehat{\mathcal{M}} \circ W_T$ (it is because the kernel is defined in real number field, the probability of being zero for every element can be ignored), the first requirement is still satisfied.

We then approximate the posteriori distribution by minimizing the KL divergence

$$KL(\mathcal{Q}(\widehat{\mathcal{M}} \circ W_T) \mid\mid \mathcal{P}(\widehat{\mathcal{M}} \circ W_T \mid \mathbb{D}), \quad \text{(S2)}$$

where $\mathcal{Q}(\cdot)$ denotes a variational distribution. We can rewrite the above KL term as

$$\log(p(\mathbb{D})) + \mathbb{E}_{\mathcal{Q}(\widehat{\mathcal{M}} \circ W_T)}[\log(\frac{\mathcal{Q}(\widehat{\mathcal{M}} \circ W_T)}{\mathcal{P}(\widehat{\mathcal{M}} \circ W_T, \mathbb{D})})]. \quad \text{(S3)}$$

---
[1] We actually broadcast (repeat) the elements in the last dimension of $\widehat{\mathcal{M}}$ to match the shape $L \times L \times 3 \times V$

Since $\log(p(\mathbb{D}))$ is the constant evidence, minimizing Eq. S2 is equivalent to minimizing the negative evidence lower bound $\mathbb{E}_{\mathcal{Q}(\widehat{\mathcal{M}} \circ W_T)}[\log(\frac{\mathcal{Q}(\widehat{\mathcal{M}} \circ W_T)}{\mathcal{P}(\widehat{\mathcal{M}} \circ W_T, \mathbb{D})})]$. This lower bound can be further rewritten as

$$KL(\mathcal{Q}(\widehat{\mathcal{M}} \circ W_T) \mid\mid \mathcal{P}(\widehat{\mathcal{M}} \circ W_T)) \\ - \sum_{t=1}^{N} \int \mathcal{Q}(\widehat{\mathcal{M}} \circ W_T) \log p(y_t \mid x_t, \widehat{\mathcal{M}} \circ W_T) \mathrm{d}\widehat{\mathcal{M}} \circ W_T. \quad \text{(S4)}$$

We assume the $\mathcal{Q}(\widehat{\mathcal{M}} \circ W_T)$ to have a factorized form and we factorize it over fusion units at each layer as

$$\prod_{l,i,u} q(\widehat{\mathcal{M}}(l, i, u) \cdot W_T(l, i, u, :)). \quad \text{(S5)}$$

We further re-parameterize the random kernel weight $W_T$ with a deterministic kernel weight multiplying a random variable that subjects to some distribution. Take a univariate Gaussian distribution $x \sim q_\theta(x) = \mathcal{N}(\mu, \sigma)$ as an example, its re-parametrization can be $x = g(\theta, \epsilon) = \mu + \sigma \epsilon$ with $\epsilon \sim \mathcal{N}(0,1)$ a parameter-free random variable, where $\mu$ and $\sigma$ are the variational parameters $\theta$. Following [3], we choose the Bernoulli distribution for the re-parametrization, which leads to

$$\widehat{\mathcal{M}}(l, i, u) \cdot W_T(l, i, u, :) = \widehat{\mathcal{M}}(l, i, u) \cdot (w_{l,i,u} \cdot z_{l,i,u}), \\ where \ z_{l,i,u} \sim Bernoulli(\widetilde{p}_{l,i,u}). \quad \text{(S6)}$$

Here $w_{l,i,u}$ is the deterministic weight matrix associated with the random weight matrix $W_T(l, i, u, :)$. Since each $\widehat{\mathcal{M}}(l, i, u)$ controls the utilization of the fusion units $u$ with binary values 1 and 0, it also subjects to Bernoulli distribution, denoted as $Bernoulli(\widehat{p}_{l,i,u})$. Therefore, we use a new Bernoulli distribution to replace the original two as

$$\widehat{\mathcal{M}}(l, i, u) \cdot W_T(l, i, u, :) = w_{l,i,u} \cdot \epsilon_{l,i,u}, \\ where \ \epsilon_{l,i,u} \sim Bernoulli(p_{l,i,u}), \quad \text{(S7)} \\ p_{l,i,u} = \widehat{p}_{l,i,u} \cdot \widetilde{p}_{l,i,u}.$$

Now, we replace the $\widehat{\mathcal{M}}(l, i, u) \cdot W_T(l, i, u, :)$ with its re-parameterized form in the second term in Eq. S4, which

leads to

$$\sum_{t=1}^{N} \int \mathcal{Q}(\widehat{\mathcal{M}} \circ W_T) \log p(y_t \mid x_t, \widehat{\mathcal{M}} \circ W_T) \mathrm{d}\widehat{\mathcal{M}} \circ W_T$$
$$= \sum_{t=1}^{N} \int p(\epsilon) \log p(y_t \mid x_t, w \cdot \epsilon) \mathrm{d}\epsilon \qquad (S8)$$
$$\approx \sum_{t=1}^{N} p(\epsilon^t) \log p(y_t \mid x_t, w \cdot \epsilon^t).$$

We use Monte Carlo estimation to approximate the integral term in the above equation, where $\epsilon^t$ indicates $t$-th sampling. Combining Eq. S4 with Eq. S8, our objective function is converted to minimizing

$$KL(\mathcal{Q}(\widehat{\mathcal{M}} \circ W_T) \parallel \mathcal{P}(\widehat{\mathcal{M}} \circ W_T))$$
$$- \sum_{t=1}^{N} p(\epsilon^t) \log p(y_t \mid x_t, w \cdot \epsilon^t). \qquad (S9)$$

The second term in Eq. S9 is equivalent to the inference of a neural network with DropPath on dataset $\{x_i, y_i\}$. However, the derivatives w.r.t. the Eq. S9 is still difficult to compute because of the KL term.

Here we leverage the Proposition 4 in [2] which proves that given fixed $M, C \in \mathbb{N}$, a probability vector $\mathbf{p} = (p_1, p_2, ..., p_C)$, and $\Sigma_h \in \mathbb{R}^{M \times M}$ diagonal positive-definite for $h = 1, 2, ..., C$, with the elements of each $\Sigma_h$ not dependent on M, and let $q(x) = \sum_{h=1}^{C} p_h \mathcal{N}(x; \mu_h, \Sigma_h)$ be a mixture of Gaussians with N components, where $\mu_h \in \mathbb{R}^M$, if assuming that $\mu_h - \mu_j \sim \mathcal{N}(0, I)$, the KL divergence between $q(x)$ and $p(x) = \mathcal{N}(0, I_k)$ can be approximated as

$$KL(q(x), p(x)) \approx \sum_{h=1}^{C} [\frac{p_h}{2}(\mu_h^t \mu_h + tr(\Sigma_h)) \qquad (S10)$$
$$- K(1 + \log 2\pi) - \log |\Sigma_h|) + p_h \log p_h].$$

Actually, Eq. S7 suggests that

$$q(\widehat{\mathcal{M}}(l, i, u) \cdot W_T(l, i, u, :))$$
$$= \delta(\widehat{\mathcal{M}}(l, i, u) \cdot W_T(l, i, u, :) - w_{l,i,u} \cdot \epsilon_{l,i,u}), \qquad (S11)$$

and we can approximate each $q(\widehat{\mathcal{M}}(l,i,u) \cdot W_T(l,i,u,:) | \epsilon_{l,i,u})$ as a narrow Gaussian with a small standard deviation $\Sigma = \sigma^2 I$. Therefore, $q(\widehat{\mathcal{M}}(l,i,u) \cdot W_T(l,i,u,:)) = \int q(\widehat{\mathcal{M}}(l,i,u) \cdot W_T(l,i,u,:) | \epsilon_{l,i,u}) p(\epsilon) d\epsilon$ is also a mixture of two Gaussians with small standard deviations (similar with the one in the above proposition), where one component fixed at zero and another one fixed at $w_{l,i,u}$. If we assume the prior of $u$ to be 'S+ST' at all layers in the template network and the prior of kernel weight to be Guassian distribution $\mathcal{N}(w_{l,i,u}; 0, I/(k_{l,i,u})^2)$, where $k_{l,i,u}$ is a prior length scale, the prior distribution of each $\widehat{M}(l,i,u) \cdot W_T(l,i,u,:)$ is still Guassian. Given the Eq. S5 and the proposition Eq. S10, it can be easily derived that

$$\frac{\partial}{\partial w \partial p} KL(\mathcal{Q}(\widehat{\mathcal{M}} \circ W_T) \parallel \mathcal{P}(\widehat{\mathcal{M}} \circ W_T))$$
$$= \frac{\partial}{\partial w \partial p} \sum_{l,i,u} KL(q(\widehat{\mathcal{M}}(l,i,u) \cdot W_T(l,i,u,:)) \parallel p(\widehat{\mathcal{M}}(l,i,u) \cdot W_T(l,i,u,:)))$$
$$\approx \frac{\partial}{\partial w \partial p} \sum_{l,i,u} \frac{(1-p_{l,i,u})k_{l,i,u}^2}{2} \|w_{l,i,u}\|^2 + p_{l,i,u} \log p_{l,i,u}. \qquad (S12)$$

In addition to the variational parameters $w_{l,i,u}$, the optimal distribution of random variable $\epsilon$ which encodes the network architecture information also needs to be found. In order to facilitate a gradient based solution, we employ Gumbel-softmax to relax the discrete Bernoulli distribution to continuous space. More specifically, instead of drawing $\epsilon_{l,i,u}$ w.r.t. the $Bernoulli(p_{l,i,u})$, we deterministically draw the $\epsilon_{l,i,u}$ with

$$\epsilon_{l,i,u} = Sigmoid(\frac{1}{\tau}[\log p_{l,i,u} - \log(1 - p_{l,i,u})$$
$$+ \log(\log r_2) - \log(\log r_1)]) \qquad (S13)$$
$$s.t.\ r_1, r_2 \sim Uniform(0,1).$$

Under this re-parametrisation, the distribution of $\epsilon_{l,i,u}$ is smooth for $\tau > 0$ and $p(\epsilon_{l,i,u}) \to Bernoulli(p_{l,i,u})$ as $\tau$ approaches 0 and. Therefore, we have well-defined gradients w.r.t. the probability $p_{l,i,u}$ by using a small $\tau$. Combining Eq. S12 and S13, we can obtain the gradients presented by the Eq. (5) in the paper.

### 1.2. Equation 7 in the paper

We use $v_0$ to denote index $(l_0, i_0, u_0)$ for simplicity. It is straightforward to derive that

$$\mathcal{P}(\widehat{\mathcal{M}}(v_0) = 0 \mid \mathbb{D})$$
$$= \int_{W_T(v_0)} \mathcal{P}(\widehat{\mathcal{M}}(v_0) = 0, W_T(v_0) \mid \mathbb{D}) \qquad (S14)$$
$$= \int_{W_T(v_0)} \mathcal{P}(\widehat{\mathcal{M}}(v_0) \cdot W_T(v_0) = 0 \mid \mathbb{D}).$$

Figure 1: Architecture of the template network used in our method. The notion 'x6' indicates that the basic module (a single grey box) is repeated by six times. The spatial stride size used for all the convolution and pooling operations is one and two, respectively. The temporal stride size is always set to be one.

Once the variational distribution $Q$ is available, it is easy to derive that

$$\begin{aligned}
&\mathcal{P}(\widehat{\mathcal{M}}(v_0) \cdot W_T(v_0) \mid \mathbb{D}) \\
&= \int_{\widehat{\mathcal{M}}(v) \cdot W_T(v), v \neq v_0} \mathcal{P}(\widehat{\mathcal{M}}(v) \cdot W_T(v) \mid \mathbb{D}) \\
&\approx \int_{\widehat{\mathcal{M}}(v) \cdot W_T(v), v \neq v_0} \prod_v q(\widehat{\mathcal{M}}(v) \cdot W_T(v)) \\
&= \int_{\widehat{\mathcal{M}}(v) \cdot W_T(v), v \neq v_0} q(\widehat{\mathcal{M}}(v_0) \cdot W_T(v_0)) \prod_{v \neq v_0} q(\widehat{\mathcal{M}}(v) \cdot W_T(v)) \\
&= q(\widehat{\mathcal{M}}(v_0) \cdot W_T(v_0)) \int_{\widehat{\mathcal{M}}(v) \cdot W_T(v), v \neq v_0} \prod_{v \neq v_0} q(\widehat{\mathcal{M}}(v) \cdot W_T(v)) \\
&= q(\widehat{\mathcal{M}}(v_0) \cdot W_T(v_0)).
\end{aligned} \quad \text{(S15)}$$

According to Eq. S7, S14 and S15, we have

$$\begin{aligned}
&\mathcal{P}(\widehat{\mathcal{M}}(v_0) = 1 \mid \mathbb{D}) \\
&= 1 - \mathcal{P}(\widehat{\mathcal{M}}(v_0) = 0 \mid \mathbb{D}) \\
&= 1 - \int_{W_T(v_0)} \mathcal{P}(\widehat{\mathcal{M}}(v_0) \cdot W_T(v_0) = 0 \mid \mathbb{D}) \\
&\approx 1 - \int_{W_T(v_0)} q(\widehat{\mathcal{M}}(v_0) \cdot W_T(v_0) = 0) \\
&= 1 - \widehat{p_{v_0}}.
\end{aligned} \quad \text{(S16)}$$

In practice we optimize $p_{l,i,u} = \widehat{p}_{l,i,u} \cdot \widetilde{p}_{l,i,u}$ as a whole, and the $\widehat{p}_{l,i,u}$ and $\widetilde{p}_{l,i,u}$ are initialized with the same value, therefore, we have

$$\widehat{p_{v_0}} = \sqrt{p_{v_0}}. \quad \text{(S17)}$$

Hereby, we have the posterior probability of fusion unit at each layer, which can be used as numerical measurements for the layer-level importance of the fusion units as described in the paper.

## 2. More Details on Experimental Setups

### 2.1. Template Network

We visualize the complete structure of the template network in Fig. 1. As can be viewed in the figure, each layer in the template network contains three convolution operations which is used to derive the three basic fusion units as illustrated in the Fig. 3 in the paper. The feature map output from each layer will be used as input for all the succeeding layers, which forms a densely-connected 3D network. Similar to [6], we also insert reduction block for memory efficiency.

### 2.2. Figure 3

Fig. 3 in the paper is a part of the whole template network (Fig. 1 in this supplementary material). It illustrates how the basic units are derived from the template network. Each grey block refers to a module that contains three convolutions as well as three drop-path operations (D1, D2, and D3). By activating each drop-path operation w.r.t. its probability, we can derive three basic fusion units, i.e., S (activate D3 only), ST (activate D2 only) and S+ST (no activation), plus one residual operation (activate D1). This design helps us to construct the probability space via network training with DropPath, as described in the Section 3.2 in the paper.

### 2.3. Basic Fusion Units

There two reasons for $U = \{S, ST, S+ST\}$ chosen instead of the full set, *i.e.*, $U = \{S, T, ST, S+T, S+ST, T+ST, S+T+ST\}$. 1) Hardware constraint. Something and Kinetics dataset both have more than 200K videos and the training time for a single trial is two weeks. If we consider all possible combinations in template network, the parameter size and training time would increase around 25% and 40%, respectively, which results in one more week for training. Besides, we have to reduce batch size from 32 to 16, which also affects the training speed and convergence badly. We can not afford such computational cost given the limited hardware resources. Therefore, we choose to only investigate the three widely investigated fusion units [25,35] in this paper. 2) Fair comparison. As discussed in Section 4.1 in the paper, we prefer that the fusion units explored in our approach are conceptually included in most of other fusion methods for fair comparison with state-of-the-arts in terms of efficiency and effectiveness.

## 3. More Results on Generalization

In order to justify that the observations obtained from the probability space can generalize, we construct new spatiotemporal strategies based on the observations discussed in Section. 4.4 in the paper, but with five very different backbone networks. They are DenseNet121[6], ResNet50[5], MobileNetV2[10], ResNeXt50[11], and ResNeXt101[11], respectively. They differ from each other in terms of topology, parameter size and FLOPs. We inflate them into 3D CNNs with the same module visualized in our template network. We compare four different fusion strategies on each backbone, i.e., optimized(Opt), fused(S+ST), spatial(S) and spatiotemporal(ST). 'Opt' means we follow the observations to design the fusion strategies (except for Densenet we directly use the best one sampled from the probability space). Please note that we can only roughly follow the observations because the network topology varies from backbone to backbone. 'S+ST' indicates that we employ S+ST convolution all the way. 'Spatial' and 'spatiotemporal' indicate using spatial convolutions and spatiotemporal convolutions all the way, respectively. Please also note that there are 3D poolings existing in the 'spatial' mode, so it is not pure 2D network. We report clip-level performance in this section for quick comparison.

On Something-Something V1 and Something-Something V2, we follow the Observation I and II to construct the strategy 'Opt' where fused convolutions are used for the first half and the last three layers of network, and spatiotemporal convolutions are applied on the remaining layers. As can be viewed from the Fig. 2(b) and 3(b), although backbones are quite different, all the networks perform better with the optimized fusion strategy. One exception is MobileNetV2, where the 'Opt' strategy is slightly worse than the fused mode. We think the reason is that MobileNetV2 is too small to fit this large dataset and any kind of bonus on the parameter size would help improve the performance greatly. The 'S+ST' mode contains 7 more 2D convolutions than the 'Opt' mode in the MobileNetV2 backbone.

Similar results can be observed on UCF101, where the 'Opt' strategy makes all the four backbone networks outperform their counterparts, which is consistent with the Observation III. Please note that the optimized strategy is equivalent to the fused strategy on UCF101 according to the Fig. 4.

On Kinetics400, we roughly follow the Observation I II and III as well as the patterns in Fig. 5(a) to use the S+ST convolutions and S convolutions periodically in the first two third of network and ST convolutions in the remaining layers for the strategy 'Opt'. We can see from the Fig. 5(b) that 'Opt' strategy still performs the best on different backbones. Due to the limited, we can not evaluate the S+ST strategy on Kinetics400 with the 3D ResNeXt101 backbone.

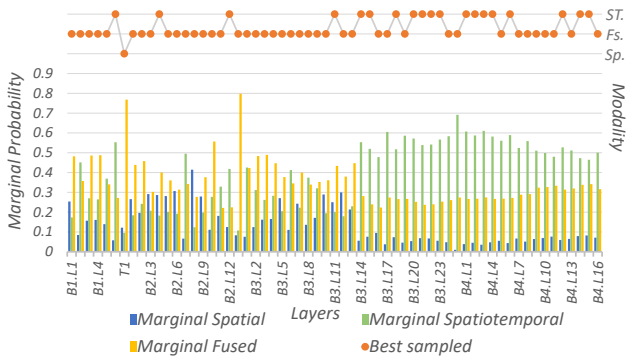

(a) Figure: Marginal probability of layer-level fusion units.

| Mod. Net. | Opt | S+ST | S | ST |
|---|---|---|---|---|
| 3D Dense.121 | 50.2 | 46.5 | 41.8 | 47.5 |
| 3D ResNet50 | 41.2 | 38.9 | 33.8 | 40.1 |
| 3D ResNeXt50 | 43.6 | 40.7 | 35.2 | 42.1 |
| 3D ResNeXt101 | 44.0 | 42.3 | 36.6 | 42.7 |

(b) Table: The performance of different backbones constructed with the spatiotemporal hints and their counter-parts.

Figure 2: Spatiotemporal fusion hints obtained from the probability space on Something-Something V1.

We also implement a small experiment to illustrate the learned probability space can help the spatiotemporal fusion strategy capture the character of the data. More specifically, we reduce the temporal resolution of input video clips by employing temporal pooling with a stride of 2 and window size of 3. We draw the corresponding marginal probability and best-sampled fusion strategy on Something V1 in Fig. 6. It can be easily observed that the layer-level preference changes accordingly. There are more spatial convolutions and less spatiotemporal convolutions utilized in the best-sampled strategy when compared with Fig. 2 in which no additional temporal pooling is used.

## 4. More Discussions on NAS

We believe that our proposed algorithm can be extended for NAS. But in this paper, it is specially designed for analyzing spatiotemporal fusion from a probabilistic view. To ensure the whole scheme is theoretically sound and correct, we have to construct a valid probability space to facilitate efficient and effective analysis with numerical measurements. To our best knowledge, no published NAS papers manage to formulate and construct such probability space.

Additionally, existing NAS methods have shortcoming in spatiotemporal analysis. For example, differentiable NAS methods such as DARTS[7] assign a scalar unit (regularized by softmax function) to each operation and jointly optimize weight and the unit in an alternative fashion. Only

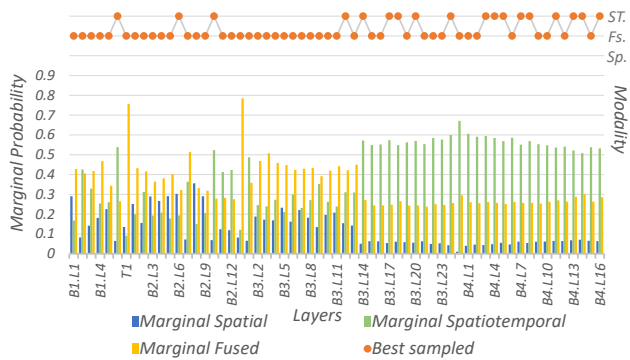

(a) Figure: Marginal probability of layer-level fusion units.

| Mod. Net. | Opt | S+ST | S | ST |
|---|---|---|---|---|
| 3D Dense.121 | 62.4 | 59.5 | 55.1 | 60.5 |
| 3D Mobile.v2 | 59.5 | 59.7 | 52.9 | 59.3 |

(b) Table: The performance of different backbones constructed with the spatiotemporal hints and their counter-parts.

Figure 3: Spatiotemporal fusion hints obtained from the probability space on Something-Something V2.

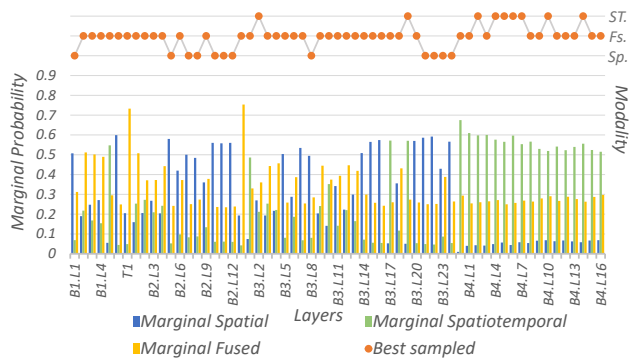

(a) Figure: Marginal probability of layer-level fusion units.

| Mod. Net. | Opt | S+ST | S | ST |
|---|---|---|---|---|
| 3D Dense.121 | 71.7 | 69.7 | 67.8 | 68.3 |
| 3D ResNeXt101 | 72.0 | - | 70.6 | 70.9 |

(b) Table: The performance of different backbones constructed with the spatiotemporal hints and their counter-parts.

Figure 5: Spatiotemporal fusion hints obtained from the probability space on Kinetics400.

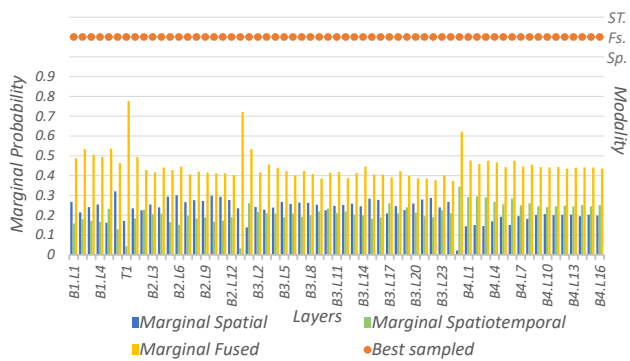

(a) Figure: Marginal probability of layer-level fusion units.

| Mod. Net. | Opt | S+ST | S | ST |
|---|---|---|---|---|
| 3D Dense.121 | 84.2 | 84.2 | 83.6 | 83.1 |
| 3D ResNet50 | 82.4 | 82.4 | 81.2 | 81.6 |
| 3D Mobile.v2 | 81.8 | 81.8 | 81.3 | 80.8 |
| 3D ResNeXt50 | 85.1 | 85.1 | 83.9 | 82.9 |

(b) Table: The performance of different backbones constructed with the spatiotemporal hints and their counter-parts.

Figure 4: Spatiotemporal fusion hints obtained from the probability space on UCF101.

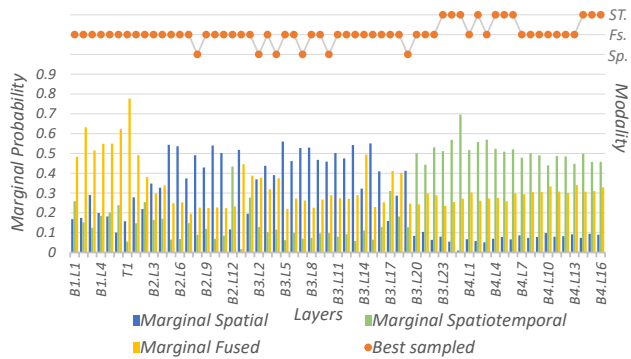

(a) Figure: Marginal probability of layer-level fusion units.

| Mod. Net. | Opt | S+ST | S | ST |
|---|---|---|---|---|
| 3D Dense.121 | 44.2 | 40.1 | 38.8 | 41.1 |

(b) Table: The performance of Densenet121 constructed with the spatiotemporal hints and their counter-parts

Figure 6: Spatiotemporal fusion hints obtained from the probability space on Something-Something V1 with less temporal resolution.

one single architecture is derived by selecting the operation that has the largest activation in each layer and the weight needs to be trained from scratch to produce final performance. It can not properly sample a group of diverse ar-

chitectures to be directly evaluated for further fusion analysis. Sampling-based NAS methods such as [1, 4] randomly sample different sub-networks and naively inherit the kernel weights from the template network. A group of candidates can be obtained in this way. However, the inherited weight may not match the sampled architectures and thus the per-

formance of each sub-network does not match the ideal performance as well, which suggests it can not be used to facilitate an effective analysis on spatiotemporal fusion in 3D CNNs. As another example, Evolution-based NAS methods iteratively derive new child architectures from population. Due to the high complexity of evolution algorithm, it cannot facilitate fine-grained layer-level spatiotemporal analysis. For instance, evolved module is repeated several times to form a homogeneous network which has few variations in terms of network-level structure diversity [8], and only connectivity among several predefined modules can be evolved [9].



\end{document}